\documentclass[10pt,journal,cspaper,compsoc]{IEEEtran}
\usepackage{times}
\usepackage{epsfig}
\usepackage{graphicx}
\usepackage{amsmath}
\usepackage{amssymb}
\usepackage{multirow}
\usepackage{color}
\usepackage{xspace}

\newcommand*{\ie}{i.e.\@\xspace}

\makeatletter
\newcommand*{\etc}{%
    \@ifnextchar{.}%
        {etc}%
        {etc.\@\xspace}%
}
\newcommand*{\etal}{%
    \@ifnextchar{.}%
        {et.al\@\xspace}%
        {et.al.\@\xspace}%
}
\makeatother

\ifCLASSOPTIONcompsoc
\else
\fi
\ifCLASSINFOpdf
\else
\fi
\begin{document}
%
\title{Collaborative Representation for Classification, Sparse or Non-sparse?}
%
%
%
%

\author{Yang~Wu,
        Vansteenberge~Jarich,
        Masayuki~Mukunoki,
        and~Michihiko~Minoh,~\IEEEmembership{Member,~IEEE}
\IEEEcompsocitemizethanks{
\IEEEcompsocthanksitem Y. Wu, M. Mukunoki, and M. Minoh are with the Academic Center for Computing and Media Studies, Kyoto University, Kyoto, 606-8501, Japan. 
E-mail: \{yangwu, mukunoki, minoh\}@mm.media.kyoto-u.ac.jp.
\IEEEcompsocthanksitem V. Jarich is with the Department of Intelligence Science and Technology, Graduate School of Informatics,
Kyoto University, Kyoto, 606-8501, Japan. 
E-mail: vansteenberge@mm.media.kyoto-u.ac.jp.}
\thanks{}}

%
%

\markboth{Preprint}
{Shell \MakeLowercase{\textit{Y. Wu, et al.}}: Collaborative Representation for Classification, Sparse or Non-sparse?}
%


\IEEEcompsoctitleabstractindextext{%
\begin{abstract}

   Sparse representation based classification (SRC) has been proved to be a simple, effective and robust solution to face recognition. As it gets popular, doubts on the necessity of enforcing sparsity starts coming up, and primary experimental results showed that simply changing the $l_1$-norm based regularization to the computationally much more efficient $l_2$-norm based non-sparse version would lead to a similar or even better performance. However, that's not always the case. Given a new classification task, it's still unclear which regularization strategy (i.e., making the coefficients sparse or non-sparse) is a better choice without trying both for comparison. In this paper, we present as far as we know the first study on solving this issue, based on plenty of diverse classification experiments. We propose a scoring function for pre-selecting the regularization strategy using only the dataset size, the feature dimensionality and a discrimination score derived from a given feature representation. Moreover, we show that when dictionary learning is taking into account, non-sparse representation has a more significant superiority to sparse representation. This work is expected to enrich our understanding of sparse/non-sparse collaborative representation for classification and motivate further research activities.

\end{abstract}

}

\maketitle

\IEEEdisplaynotcompsoctitleabstractindextext

 \ifCLASSOPTIONpeerreview
 \begin{center} \bfseries EDICS Category: 3-BBND \end{center}
 \fi
%
\IEEEpeerreviewmaketitle

\section{Introduction}


Recently, a simple approach called sparse representation based classification (SRC) \cite{SRC_TPAMI09}, has shown quite impressive results on face recognition and also some other classification tasks \cite{SR4CVPR_PIEEE10}. It minimizes the $l_2$-norm based error on reconstructing a test sample with a linear combination of all the training samples whilst limiting the sparsity of reconstruction coefficients. The sparsity term tends to force larger coefficients to be assigned to training samples in the same class as which the test sample belongs to, making such coefficients discriminative for classification. Since the ideal $l_0$-norm for modeling sparsity leads to a computationally expensive or even infeasible combinatorial optimization problem, SRC adopts $l_1$-norm to approximate the $l_0$-norm, though it is still a bit time consuming due to its unavoidable iterative optimization. The main weakness of SRC is that it has two preconditions for ensuring a good performance \cite{SR4CVPR_PIEEE10}: the training samples need to be carefully controlled and the number of samples per class has to be sufficiently large, and there is a lack of quantitative criteria for verifying whether they are satisfied or not.

Later research argued that SRC's success lies in the collaborative representation using all the training samples, but not the $l_1$-norm based regularization which makes the representation coefficients sparse \cite{CRC_ICCV11}. It has shown that the $l_1$-norm can be replaced by the computationally much more efficient $l_2$-norm, without sacrificing the performance. Therefore, for a better understanding and comparison, these two models were both treated as collaborative representation based classification (CRC), and the regularization term was used to differentiate them. Here we follow the same notation and name them CRC$\_l_1$ and CRC$\_l_2$, standing for sparse representation and non-sparse representation, respectively, because the representation coefficients regularized by $l_1$-norm are widely-regarded to be sparse while the ones regularized by $l_2$-norm are generally non-sparse.

Since the birth of CRC$\_l_2$, more and more attention has been paid to $l_2$-norm based regularization for collaborative representation due to its attractive effectiveness and efficiency. Though experiments on both CRC$\_l_2$ itself \cite{CRC_CoRR12} and its extensions \cite{ECRC_IEICETIS13} have shown their superiority to the sparse representation competitors, there are counterexamples reported as well \cite{CRC_CoRR12, WCRC_PRL13}. The uncertain relative superiority between sparse and non-sparse CRC models confuses people who have limited research experiences on them, and there is still a lack of an in-depth and reliable criterion for preselecting the more promising model for a given task. This paper presents our study for solving this problem. Specifically, we contribute in two aspects:
\begin{itemize}
  \item We propose an analytic scoring function, depending on only the given feature vectors and the size of the dataset, for predicting whether the CRC model should be sparse or non-sparse. Extensive and representative experiments on various classification tasks have demonstrated the effectiveness of this function.

  \item We further discuss the important direction of extending collaborative representation with dictionary learning whilst proposing a very simple dictionary learning approach for non-sparse collaborative representation (named DL-NSCR), which is as far as we are aware the first of its kind. Extensive experiments have shown that DL-NSCR is generally superior to the most similar as well as other state-of-the-art dictionary learning models for sparse representation in term of effectiveness, robustness and efficiency.
\end{itemize}

The rest parts of the paper are organized as follows. A brief introduction of the background knowledge for collaborative representation is given in section \ref{sec:background_knowledge} after commenting on the related work in section \ref{sec:related_work}. Section \ref{sec:our_proposal} presents the details of DL-NSCR. All the experiments and results are stated and analyzed in section \ref{sec:experiments}, while the conclusions and future work are given in section \ref{sec:conclusions}.

\section{Related work}
\label{sec:related_work}

\subsection{Comparison between CRC$\_l_1$ and CRC$\_l_2$}

After proposing CRC$\_l_2$, Zhang \etal \cite{CRC_CoRR12} have done more experiments on comparing CRC$\_l_1$ with CRC$\_l_2$, along with their robust versions for handling occlusions/corruptions. They concluded that the relative superiority between them depends on the feature dimensionality of data. It was supposed that high dimensionality corresponds to high discrimination power and in this case the coefficients tend to be naturally and passively sparse even without sparse regularization, so CRC$\_l_2$ can do a better job than CRC$\_l_1$. If the dimension is very low, it will lead to the opposite result. While we agree with the point that CRC$\_l_2$ favors higher dimension, we don't think the relative superiority only depends on feature dimensionality. Note that the assumption of correspondence between feature dimensionality and data discrimination power in Zhang \etal's work \cite{CRC_CoRR12} is unreliable, because you may have arbitrarily different features with quite different discrimination abilities for a given dimensionality. Even when the same feature vectors are projected into different spaces using certain dimension reduction approach, there are still counterexamples to the effectiveness of using dimensionality as the indicator, as shown in \cite{WCRC_PRL13}. Besides this issue, there is another drawback of Zhang \etal's  experiments: they are limited to only 3 face datasets, though different subsets/variations of them have been tested on.



\subsection{Comparison on extended CRC models}

The debate between sparse and non-sparse collaborative representation is not only limited to the simplest CRC$\_l_1$ and CRC$\_l_2$ models. Recently, a model called Extended SRC (ESRC) \cite{ESRC_TPAMI12} added another generic dictionary based on a third-party dataset for handling the within-class variations. Since the additional dictionary may be able to cover the possibly large changes between the test sample and the corresponding training samples from the same class, ESRC can be applied to single-shot recognition problems where only a single training sample is available for each class. Later on, a non-sparse version Extended CRC (ECRC) \cite{ECRC_IEICETIS13} was published. The only difference between ECRC and ESRC is that ECRC uses $l_2$-norm instead of $l_1$-norm for coefficients regularization. Experimental results on several widely used face recognition datasets showed that ECRC is much faster and more effective than ESRC. Though being interesting and valuable, the comparison has a limitation that both of these two models depend on the third-party data which brings two new problems: the relative superiority between ESRC and ECRC may depend on the third-party data and the selection and evaluation of this third-party data is not a trivial task.

\subsection{An important unexplored area: dictionary learning}

There is another more important and influential direction for enhancing the collaborative representation models --- dictionary learning (DL), i.e., learning a discriminative dictionary from the training data instead of directly using it as the dictionary. Generally speaking, dictionary learning can significantly improve the discrimination and generalization abilities of CRC models without relying on any other additional data.

Quite a few publications on DL can be found for sparse representation, but as far as we are aware no such model has ever been proposed for non-sparse representation. The existing DL models can be roughly grouped into three categories: making the dictionary discriminative, learning a discriminative model for classification with the coefficients, and simultaneously optimizing the dictionary and the coefficients-based classification model.

The first group \cite{Meta-face_ICIP10, DLSI_CVPR10} follow the classification model of CRC$\_l_1$ in using only the class-specific reconstruction error, directly targeting a discriminative dictionary. The second group try to learn discriminative classification models on the sparse representation coefficients, including logistic regression \cite{SupervisedDL_NIPS08} and linear regression (D-KSVD \cite{D-KSVD_CVPR10} and LC-KSVD \cite{LC-KSVD_CVPR11} which adds one more linear regression term to D-KSVD to further enhance the label consistency within each class). The third group so far contain only one representative work called Fisher discrimination dictionary learning (FDDL) \cite{FDDL_ICCV11}. It has a discriminative fidelity term which minimizes the reconstruction error using both a global dictionary and class-specific sub-dictionaries, while at the same time minimizes the ability of the sub-dictionaries on reconstructing samples from different classes. Besides that, FDDL also has a discriminative coefficient term which utilizes the Fisher discriminant to make the coefficients discriminative. Very recently, a new model called DL-COPAR \cite{DL-COPAR_ECCV12} developed the idea proposed in DLSI \cite{DLSI_CVPR10} on exploring the common bases of sub-dictionaries by explicitly separating the particularity (class-specific sub-dictionaries) and commonality (a common sub-dictionary) in dictionary learning. Meanwhile, it also inherited the incoherence term from \cite{DLSI_CVPR10} and the third part of the fidelity term in FDDL to make the class-specific sub-dictionaries as discriminative as possible.


Despite their differences in learning the discriminative model, all these approaches enforce the sparsity of the coefficients using either $l_0$-norm or $l_1$-norm, which is usually computationally expensive.

\section{Collaborative representation and dictionary learning}
\label{sec:background_knowledge}

\subsection{Sparse representation}
Suppose a training dataset $X = [X_1, \dots, X_L ] \in \mathbb{R}^{d\times n}$ is given, where $n$ denotes the total number of samples; $d$ denotes their feature dimension; $L$ is the number of classes; and $X_i$ denotes the $n_i$ samples belonging to class $i$. SRC (i.e., CRC$\_l_1$) seeks a reconstruction of test sample ${\bf{y}} \in \mathbb{R}^d$ via a linear combination of all the training samples $X$, while at the same time minimizes the $l_1$-norm sparsity of the reconstruction coefficients. The coding model can be formulated as:
\begin{equation}\label{eq:SRC_coding}
\hat \alpha  = \arg {\min _\alpha }\;\left\| {{\bf{y}} - X\alpha } \right\|_2^2{\rm{ + }}{\lambda _{\rm{1}}}{\left\| \alpha  \right\|_1^2},
\end{equation}
where $\hat \alpha  = [{\hat \alpha }_1, \dots, {\hat \alpha }_L]$ are the concatenated reconstruction coefficients corresponding to training samples of different classes, and $\lambda _{\rm{1}}$ is a regulatory parameter for weighting the regularization of $\alpha$. For classification, CRC$\_l_1$ computes the representation residual for each class: 
\begin{equation}\label{eq:SRC_residuals}
{r_i}\left( {\bf{y}} \right) = {\left\| {{\bf{y}} - {X_i}{{\hat \alpha }_i}} \right\|^2_2},\forall i \in \left\{ {1, \ldots ,L} \right\},
\end{equation}
and then classifies ${\bf{y}}$ by $C\left( {\bf{y}} \right) = \arg {\min _i}{r_i}\left( {\bf{y}} \right)$.

\subsection{Non-sparse representation}

According to the recent arguments from \cite{CRC_ICCV11}, in the SRC (i.e., CRC$\_l_1$) model, it is the collaborative representation with all classes but not the $l_1$-norm regularization term that truly contributes to the good face recognition performance. Therefore, they proposed the following non-sparse scheme (CRC$\_l_2$ in this paper) which replaces Equation \ref{eq:SRC_coding} by
\begin{equation}\label{eq:CRC_coding}
\hat \alpha  = \arg {\min _\alpha }\;\left\| {{\bf{y}} - X\alpha } \right\|_2^2{\rm{ + }}{\lambda _{\rm{1}}}{\left\| \alpha  \right\|_2^2}.
\end{equation}

The biggest benefit of this replacement is the dramatic reduction of computational cost, because this new convex optimization problem has a closed-form solution
\begin{equation}\label{eq:CRC_solution}
\hat \alpha  = {\left( {X^T{X} + \lambda_{\rm{1}}  \cdot I} \right)^{ - 1}}X^T {\bf{y}}.
\end{equation}
More attractively, this solution is just a linear projection of ${\bf{y}}$, and the projector $P = {\left( {X^T{X} + \lambda_{\rm{1}}  \cdot I} \right)^{ - 1}}X^T$ is independent of ${\bf{y}}$. $P$ can be pre-computed given the training data, so it doesn't require a separate optimization process for each test sample as CRC$\_l_1$ demands.

Though the $l_2$-norm based regularizer in Equation \ref{eq:CRC_coding} is no longer a sparsity constraint on the coefficients, it still has the potential to induce the competition among training samples from all candidate classes, which may cause the right class to have relatively smaller reconstruction error and larger $l_2$-norm values of coefficients. Therefore, CRC$\_l_2$ computes the normalized residuals
\begin{equation}\label{eq:CRC_residuals}
{r_i}\left( {\bf{y}} \right) = \left\| {{\bf{y}} - {X_i}{{\hat \alpha }_i}} \right\|_2^2/{\left\| {{{\hat \alpha }_i}} \right\|_2}, \forall i,
\end{equation}
and then classifies ${\bf{y}}$ by $C\left( {\bf{y}} \right) = \arg {\min _i}{r_i}\left( {\bf{y}} \right)$.

\subsection{Dictionary learning}
\label{subsec:DL-SCR}

Instead of using the training data itself as the reconstruction dictionary, dictionary learning techniques seek to learn a compact yet over-complete dictionary from the training data, so that it can scale up to large amounts of training samples while at the same time being as discriminative as possible. A general dictionary learning model is:
\begin{equation}\label{eq:DL_objective}
\begin{array}{l}
\left\langle {{D^*},{W^*},{A^*}} \right\rangle \\
 = \arg \mathop {\min }\limits_{D,W,A} \left\{ {r\left( {X,D,A} \right) + {\lambda _1}\left\| A \right\|_p^2 + {\lambda _2}f\left( W,A \right)} \right\},
\end{array}
\end{equation}
where $D = [D_1, \dots, D_L ] \in \mathbb{R}^{d\times K}$ is the learned dictionary from $X$ (usually $K\leq n$); $A$ denotes the reconstruction coefficients over $D$ for $X$; $W$ denotes the learned parameters of the discriminative model $f(W,A)$ for classification with $A$; $r(X,D,A)$ is the discriminative reconstruction model defined over $D$ (called the discriminative fidelity in \cite{FDDL_ICCV11}); and $\lambda _1$ and $\lambda _2$ are trade-off parameters.

Though most of the existing dictionary learning models can be covered by the above general model (some models do not have the $f(W,A)$ term), they vary in their detailed design of $r(X,D,A)$ and $f(W,A)$, resulting in different performances and speeds. However, all the proposed dictionary learning models have to iteratively optimize the model parameters $D$ (and $W$) and the coefficients $A$ due to the difficulty on optimizing them simultaneously. As mentioned before, so far only $p=0$ and $p=1$ have been explored. 

\section{Dictionary learning for non-sparse representation}
\label{sec:our_proposal}

The proposed DL-NSCR model inherits the design of the $r(X,D,A)$ term from FDDL \cite{FDDL_ICCV11}, but discards its time-consuming $f(W,A)$ term to keep the model light. Note that there is a big difference between it and other existing dictionary learning models: it adopts the computationally efficient $l_2$-norm ($\| \cdot \|_F$ in matrix format) to regularize $A$.

\subsection{Learning model}

In DL-NSCR, $r(X,D,A)$ is designed as follows.
\begin{align}\label{eq:DL-NSCR_dictionary_term}
r\left( {X,D,A} \right) & =  \left\| {X - DA} \right\|_F^2 + \sum\limits_{i = 1}^L \left\| {{X_i} - {D_i}A_i^i} \right\|_F^2 \nonumber   \\
  & + \sum\limits_{i = 1}^L {\sum\limits_{j = 1,j \ne i}^L {\left\| {{D_i}A_j^i} \right\|_F^2} },
\end{align}
where $\| \cdot \|_F$ denotes the Frobenius norm \footnote{Frobenius norm is a generalization of $l_2$-norm (squared root of sum of squares) from dealing with vectors to operating on matrices.}, and $A_j^i$ denotes the coefficients corresponding to the sub-dictionary $D_i$ of class $i$ for those samples from $X_j$ (\ie the columns of $A_j^i$ correspond to class $j$). In this model, the first term is the \emph{overall reconstruction error} (ensuring that $D$ can well represent $X$); the second term is the \emph{class-specific reconstruction error} (forcing $D_i$ to be able to well represent $X_i$); and the third term is the \emph{confusion factor} (restricting $D_i$'s ability on reconstructing samples from any other class rather than $i$). It is easy to tell that such a definition of $r(X,D,A)$ will force $D$ to be discriminative.

With the term of $r(X,D,A)$, the overall learning model of DL-NSCR is as simple as
\begin{equation}\label{eq:DL-NSCR_objective}
\left\langle {{D^*},{A^*}} \right\rangle  = \arg \mathop {\min }\limits_{D,A} \left\{ {r\left( {X,D,A} \right) + {\lambda _1}\left\| A \right\|_F^2} \right\}.
\end{equation}

\subsection{Optimization}

Similar to other dictionary learning algorithms, the optimization of DL-NSCR is done iteratively between optimizing A and optimizing D until the iteration converges.

\subsubsection{Initialization}

We use principle component analysis for initializing $D_i$ with samples from class $i$. However, it is also acceptable to initialize with random numbers, which will only cost a few more iterations.

\subsubsection{Optimizing A given a fixed D}

Given $D$, thanks to the fact that Frobenius norm is decomposable, optimizing $A$ is equivalent to optimizing $A_i$ for each $i\in \{1, \dots, L\}$ independently as follows.
\begin{equation}\label{eq:DL-NSCR_optimize_Ai}
A_i^* = \arg \mathop {\min }\limits_{{A_i}} \left\{ \begin{array}{l}
\left\| {{X_i} - D{A_i}} \right\|_F^2 + \left\| {D{S_{\backslash i}}S_{\backslash i}^T{A_i}} \right\|_F^2\\
 + \left\| {{X_i} - D{S_i}S_i^T{A_i}} \right\|_F^2 + {\lambda _1}\left\| {{A_i}} \right\|_F^2
\end{array} \right\},
\end{equation}
where
\begin{equation}\label{eq:DL-NSCR_optimize_Ai_Si}
\begin{array}{l}
{S_i} = \left[ {\begin{array}{*{20}{c}}
{{{\rm O}_{\sum\nolimits_{m = 1}^{i - 1} {{K_m}}  \times {K_i}}}}\\
{{I_{{K_i} \times {K_i}}}}\\
{{{\rm O}_{\sum\nolimits_{m = i + 1}^L {{K_m}}  \times {K_i}}}}
\end{array}} \right],\\
{S_{\backslash i}} = \left[ {{S_1}, \cdots ,{S_{i - 1}},{S_{i + 1}}, \cdots ,{S_L}} \right],
\end{array}
\end{equation}
with $K_i, i\in \{1,\dots,L\}$ denoting the dictionary size of $D_i$. Here $S_i$ and $S_{\backslash i}$ are matrices for selecting the specific sub-dictionaries, while ${\rm O}$ and $I$ denote zero matrix and identity matrix, respectively. This optimization problem can be rewritten into a simpler form
\begin{equation}\label{eq:DL-NSCR_optimize_AiS}
A_i^* = \arg \mathop {\min }\limits_{{A_i}} \left\{ {\left\| {{R_i} - {Z_i}{A_i}} \right\|_F^2 + {\lambda _1}\left\| {{A_i}} \right\|_F^2} \right\},
\end{equation}
where
\begin{equation}\label{eq:DL-NSCR_optimize_AiS_YZ}
{R_i} = \left[ {\begin{array}{*{20}{c}}
{{X_i}}\\
{{X_i}}\\
{{{\rm O}_{d \times {n_i}}}}
\end{array}} \right],{\rm{  }}{Z_i} = \left[ {\begin{array}{*{20}{c}}
D\\
{D{S_i}S_i^T}\\
{D{S_{\backslash i}}S_{\backslash i}^T}
\end{array}} \right].
\end{equation}

With only the Frobenius norm in its optimization objective function, $A_i$ has a closed-form solution
\begin{equation}\label{eq:DL-NSCR_optimize_Di_solution}
A_i^* = {\left( {Z_i^T{Z_i} + {\lambda _1} \cdot I} \right)^{ - 1}}Z_i^T{R_i}.
\end{equation}
Equation \ref{eq:DL-NSCR_optimize_AiS} has exactly the same form as the CRC$\_l_2$ model, so it is computationally very efficient. 

\subsubsection{Optimizing D given a fixed A}

When fixing $A$, the term ${\lambda _1}\left\| A \right\|_F^2$ becomes a constant, however, $D$ is still impossible to be optimized as a whole because the objective function in Equation \ref{eq:DL-NSCR_objective} has two terms which are functions of sub-dictionaries $D_i, i\in \{1, \dots, L\}$ but not the overall dictionary $D$. Therefore, we optimize $D_i$s one-by-one, assuming the others are fixed. Concretely,
\begin{equation}\label{eq:DL-NSCR_optimize_Di}
D_i^* = \arg \mathop {\min }\limits_{{D_i}} \left\{ {\left\| {{U_i} - {D_i}{V_i}} \right\|_F^2 } \right\},
\end{equation}
where
\begin{equation}\label{eq:DL-NSCR_optimize_Di_UV}
\begin{array}{l}
{U_i} = \left[ {X - {D_{\backslash i}}{A^{\backslash i}},{X_i},{{\rm O}_{d \times \left( {n - {n_i}} \right)}}} \right],\\
{V_i} = \left[ {{A^i}, A_i^i, A_{\backslash i}^i} \right].
\end{array}
\end{equation}
In Equation \ref{eq:DL-NSCR_optimize_Di_UV}, $D_{\backslash i}$ denotes all the $D_j$s with $j\neq i$ and $A^{\backslash i}$ denotes the corresponding coefficients (\ie without $A^i$), where ``$\backslash i$'' means without class $i$. ${\rm O}_{d \times \left( {n - {n_i}} \right)}$ denotes a $d \times \left( {n - {n_i}} \right)$ dimensional zero matrix, where $n_i$ is the number of samples in class $i$ and $n = \sum\nolimits_{i = 1}^L {{n_i}}$.

Equation \ref{eq:DL-NSCR_optimize_Di} has a closed-form solution
\begin{equation}\label{eq:DL-NSCR_optimize_Di_solution}
D_i^* = {U_i}V_i^T{\left( {{V_i}V_i^T } \right)^{ - 1}}.
\end{equation}
Note that optimizing $D_i$ depends on a given $D_{\backslash i}$, which means once $D_i$ is updated, it should be used to update each $D_j, j\neq i$ in $D_{\backslash i}$. This is a chicken-and-egg problem, so a straightforward solution consists in updating all the $D_i$s iteratively until convergence. However, since we are iterating between optimizing $A$ and updating $D$, a converged $D$ will soon been changed once $A$ is recomputed. Therefore, in our implementation we ignored the inner-iteration in $D$'s optimization, and found it still worked quite well.

\subsection{Classification model}

After learning $D$, we can use it for solving both single-sample based and set based classification problems, all of which will be covered in our experiments.

For single-sample based classification, we follow CRC on reconstructing an arbitrary test sample ${\bf{y}}$ by $D$ with its reconstruction coefficients obtained by solving
\begin{equation}\label{eq:DL-NSCR_coding}
\hat \alpha  = \arg {\min _\alpha }\;\left\| {{\bf{y}} - D\alpha } \right\|_2^2{\rm{ + }}{\lambda _{\rm{1}}}{\left\| \alpha  \right\|_2},
\end{equation}
whose closed-form solution is $\hat \alpha  = {\left( {D^T{D} + \lambda_{\rm{1}}  \cdot I} \right)^{ - 1}}D^T {\bf{y}}$. Then ${\bf{y}}$ is classified by $C\left( {\bf{y}} \right) = \arg {\min _i}{r_i}\left( {\bf{y}} \right)$ with
\begin{equation}\label{eq:DL-NSCR_classification}
{r_i}\left( {\bf{y}} \right) = \left\| {{\bf{y}} - {D_i}{{\hat \alpha }_i}} \right\|_2^2/{\left\| {{{\hat \alpha }_i}} \right\|_2}.
\end{equation}

For set based classification, we have a similar reconstruction model which just replaces ${\bf{y}}$ by a set of test samples $Y$, and the reconstruction coefficients for $Y$ are $\hat A  = {\left( {D^T{D} + \lambda_{\rm{1}}  \cdot I} \right)^{ - 1}}D^T Y$. Then, we classify $Y$ by $C\left( Y \right) = \arg {\min _i}{r_i}\left( Y \right)$ with
\begin{equation}\label{eq:DL-NSCR_classification_set}
r_i \left( Y \right)  =  \left\| {Y - D_i A^i} \right\|_F^2
  + \sum\limits_{j = 1,j \ne i}^L {\left\| {{D_i}A_j^i} \right\|_F^2}.
\end{equation}

\section{Experiments and results}
\label{sec:experiments}

We conduct our experiments on five visual recognition tasks using nine public benchmark datasets. They are chosen to cover different scenarios: appearance-based face recognition in controlled environment, texture recognition focusing on texture information, leaf categorization using shape information, food categorization with the stuffs rich of highly varying color, texture and shape information, and appearance-based across-camera person re-identification in uncontrolled environment. Besides that, we follow \cite{i-LIDS-MAAA,CSA_AVSS12} on varying the number of samples per class for 2 of the 3 adopted re-identification datasets with 3 different values for each of them to investigate how this factor influences the performance of the tested models. Therefore, there are totally 13 different data settings for experiments.

Such extensive and diverse experiments extend the scope of collaborative representation from face recognition to general recognition/classification tasks with various statistics, properties, and feature representations. They are different from many other experiments in the literature which use artificially generated versions of the same dataset (for example, changing the feature dimension by applying dimension reduction methods), because the artificial data may override the true factors we're looking for.






For all the methods to be compared in each experiment, we used exactly the same features and data splits (10 splits for averaging if allowed). The regulatory parameter for the sparse/non-sparse regularization term in all the concerned models are set to be the same for a fair comparison. Specifically, we have it set to 0.5 for person re-identification datasets and 0.0001 for all the others, which was proved to be a good choice. We notice that finding the best value for this regulatory parameter and analyzing the sensitivity of it for each method are important open issues, but they are not the focus of this paper.

\subsection{Datasets and settings}

  \noindent \textbf{Face Recognition.} For face recognition, we choose two widely-used datasets: Extended Yale B \cite{ExtendedYB_TPAMI01} and AR \cite{AR_CVCTR98}. The Extended Yale B dataset contains 2414 frontal-face images belonging to 38 individuals. These images were captured under different lighting conditions with various facial expressions. We preprocessed the data according to \cite{SRC_TPAMI09}: using cropped images with a size of $198\times 168$ pixels; randomly selecting half of the samples for training and testing (about 32 samples per person); projecting each image into a 504-dimensional feature space using a random matrix generated from a zero-mean normal distribution with its rows normalized (by $l_2$-norm). The size of the class-specific dictionary $K_i$ for all the dictionary learning models was set to 15, so $K=570$, as suggested by \cite{LC-KSVD_CVPR11}. For DL-COPAR, which will be compared with, we had $K_i=14$ and the common sub-dictionary size set to be 38. The AR dataset contains more variations than the Extended Yale B dataset, including illumination, expression and disguises changes. Following \cite{SRC_TPAMI09} and \cite{FDDL_ICCV11}, we use a subset of 100 subjects (50 male and 50 female) with only 14 images containing illumination and expression changes for each of them (7 from session 1 for training and another 7 from session 2 for testing). We use 300-dimensional Eigenfaces for feature representation. $K_i$ was set to 7. For DL-COPAR, $K_i=6$ and the common sub-dictionary is assigned a size of 2. Since the training data and the test data are all fixed, only one round of experiment was conducted.

  \noindent \textbf{Texture Recognition.} We work on two representative datasets: the KTH-TIPS dataset and the CUReT dataset, because both of them have many samples for each class (may satisfy SRC's first precondition) and great within-class variations including illumination, viewpoint and scale changes. They are also different from each other in the sense that KTH-TIPS has greater within-class variations while CUReT has significantly more classes. We adopt the 1180-dimensional PRI-CoLBP$_0$ feature descriptor proposed in \cite{PRI-CoLBP_ECCV12} due to its high performances. $K_i$ was chosen to be 10 for both datasets, and the suggested common sub-dictionary size for DL-COPAR is 5.

  \noindent \textbf{Leaf Categorization.} The popular Swedish leaf dataset is used here for leaf recognition. It was carefully built with all the leaves well-aligned, open, flat, and complete. Only one side of each leaf is photographed against a clean background. Though its strict settings make the problem much easier than it might be in real applications, it has the advantage of making the problem clean and focused, i.e., distinguishing different leaf species mainly by their flat shapes. This dataset contains 15 species of leaves, with 75 images for each of them. Following the state-of-the-art model spatial PACT \cite{Spatial_PACT_PAMI11}, we have 25 images per class sampled for training and the rest left for testing. $K_i$ was set to 12, and the common sub-dictionary size for DL-COPAR was set to 2 (which was better than others). Again, we use PRI-CoLBP$_0$ as the feature descriptor.

  \noindent \textbf{Food Categorization.} It is a relatively less popular visual recognition problem due to its difficulty and the lack of good benchmark datasets. The recently released Pittsburgh Food Image Dataset(PFID) might be a good starting point. It owns fast food images and videos collected from 13 chain restaurants and acquired under lab and realistic settings. Following Yang \etal. \cite{Food_Yang_CVPR10}, we focus on the same set of 61 categories of specific food items with background removed. This set has three different instances for each category, which were bought from different chain stores on different days. Each instance has six images taken from different viewpoints. We follow the standard experimental protocol in using 12 images from two instances for training and the other 6 images from the third instance for testing. This allows a 3-fold cross-validation. The 1180-dimensional PRI-CoLBP$_0$ feature is extracted for each color channel and thus the whole feature vector is 3540-dimensional. We set $K_i=6$ for all dictionary learning models and the common sub-dictionary for DL-COPAR had a size of 3.

  \noindent \textbf{Person re-identification.} We experiment on three recently built datasets ``iLIDS-MA'', ``iLIDS-AA'' \cite{i-LIDS-MAAA} and CAVIAR4REID\cite{CAVIAR4REID_BMVC11}. They are representatives of cross-camera re-identification with non-overlapping views in real scenarios. The first two were collected from the i-LIDS video surveillance data captured at an airport, while the third one consists of several sequences filmed in a shopping centre. The iLIDS-MA dataset has 40 persons with exactly 46 manually cropped images per camera for each person, resulting in 3680 images in total. Unlike iLIDS-MA, the iLIDS-AA dataset was extracted automatically using a HOG-based detector instead of manual annotation. Such a property simulates the localization errors of human detection and tracking in real systems. Moreover, iLIDS-AA is also much bigger than iLIDS-MA. It contains as many as 100 individuals totaling 10754 images. Since it was automatically annotated, the number of images for each person varies from 21 to 243.

  For both datasets, a certain number of samples (10, 23, or 46) are randomly sampled from one camera (Camera 3) for each person to serve as the training data (i.e. the gallery), and the same amount of samples are randomly sampled from the other camera (Camera 1) for testing (i.e. as queries). It results in three different versions for each dataset, named as ``iLIDS-MA10'', ``iLIDS-MA23'', etc. Note that for iLIDS-MA46 there is only one data sampling result. Compared with iLIDS-MA and iLIDS-AA datasets, CAVIAR4REID has broader resolution changes and larger pose variations. We follow \cite{CAVIAR4REID_BMVC11} on training with 22 specified subjects and testing on the other 50 subjects. Each set (either for gallery or query) contains 5 randomly sampled images. Like \cite{CSA_AVSS12}, we perform multiple-shot re-identification and treat it as a set-based classification problem. Therefore, the set-based classification model of DL-NSCR is used.

  The person re-identification problem is usually treated as a ranking problem, and it is desired that the correct match for a given querying individual appears in the top-ranked candidates, so we used the cumulative recognition rate at rank top 10\% as the effectiveness measure. We used exactly the same 400-dimensional color and texture histograms based features as adopted in \cite{TPCR_ICPR12} for all the methods. We had $K_i, \forall i$ chosen to be the same as the number of samples per class for the iLIDS-MA dataset and the CAVIAR4REID dataset, but for the iLIDS-AA dataset they were set to be $K_i=8, \forall i$. The common sub-dictionary size for DL-COPAR got the same value as $K_i$ for all the datasets.

\setlength{\tabcolsep}{3.2pt}
\begin{table*}[ht]
\caption{Statistics of the datasets and prediction criteria for inferring when shall collaborative representation be sparse or non-sparse, in comparison with the actual performances. $d$, $C$, $n_i$, $n$ denote the feature dimensionality, number of classes, number of training samples per class, and the total number of training samples, respectively. The four special datasets (see the text for explanation) are marked with a star.}
\centering{ \footnotesize
\begin{tabular}{lrrrrrrrrrrrrrrr}
\hline\\[-2ex]
   & &\multicolumn{4}{c}{\textbf{Statistics}} &   &\multicolumn{5}{c}{\textbf{``Non-sparse vs. Sparse'' Prediction}} & &\multicolumn{3}{c}{\textbf{Actual Performance}}\\
              \cline{3-6} \cline{8-12} \cline{14-16}
   \textbf{Dataset} &   & $d$ & $C$ & $n_i$ & $n$ & & MPD* & FDR & FDR$\times d$ & FDR$/n$ & FDR$\times d/n$ &  & CRC$\_l_1$ & CRC$\_l_2$ & ERR \\
\hline \\[-2ex]
Extended Yale B& & 504 & 38 &$\sim$32& 1207 & & 0.653 & 24.81 & 12506 & 0.0206 & 10.36 & & 0.951 &\textbf{0.976}&0.510\\
\emph{AR*}     & & 300 & 100&   7    & 700  & & 0.714 & 71.40 & 21420 & 0.1020 & 30.60 & & 0.898 &\textbf{0.919}&0.206\\
KTH-TIPS       & & 1180& 10 &   40   & 400  & & 0.920 &  9.20 & 10856 & 0.0230 & 27.14 & & 0.957 &\textbf{0.971}&0.326\\
CUReT          & & 1180& 61 &   46   & 2806 & & 0.936 & 57.10 & 67373 & 0.0203 & 24.01 & & 0.824 &\textbf{0.933}&0.619\\
Swedish Leaf   & & 1180& 15 &   25   & 375  & & 0.923 & 13.85 & 16337 & 0.0369 & 43.57 & & 0.958 &\textbf{0.991}&0.786\\
\emph{Food*}   & & 3540& 61 &   12   & 732  & & 0.220 & 13.42 & 47506 & 0.0183 & 64.90 & & 0.311 &\textbf{0.349}&0.055\\
iLIDS-MA10     & & 400 & 40 &   10   & 400  & & 0.500 & 20.00 &  8000 & 0.0500 & 20.00 & & 0.750 &\textbf{0.778}&0.112\\
iLIDS-MA23     & & 400 & 40 &   23   & 920  & & 0.593 & 23.72 &  9488 & 0.0258 & 10.31 & & 0.785 &\textbf{0.790}&0.023\\
\emph{iLIDS-MA46*}& &400&40 &   46   & 1840 & & 0.625 & 25.00 & 10000 & 0.0136 &  5.44 & & 0.800 &\textbf{0.825}&0.125\\
iLIDS-AA10     & & 400 & 100&   10   & 1000 & & 0.238 & 23.80 &  9520 & 0.0238 &  9.52 & & 0.658 &\textbf{0.677}&0.056\\
iLIDS-AA23     & & 400 & 100&$\sim$23&$\sim$2300& &0.270&27.00& 10800 & 0.0117 &  4.70 & & \textbf{0.731} &0.690&-0.150\\
iLIDS-AA46     & & 400 & 100&$\sim$46&$\sim$4600& &0.351&35.10& 14040 & 0.0076 &  3.05 & & \textbf{0.779} &0.667&-0.510\\
\emph{CAVIAR4REID*}& &400&50&   5    & 250  & & 0.190 &  9.50 &  3800 & 0.0380 & 15.20 & & 0.478 &\textbf{0.514}&0.069\\
\hline
\end{tabular}}
\label{table:Sparse_VS_NonSparse}
\end{table*}

\subsection{Sparse or non-sparse representation?}

For justifying whether collaborative representation should be sparse or non-sparse, we compare the $l_1$-norm regularization with the $l_2$-norm regularization, i.e., compare CRC$\_l_1$ with CRC$\_l_2$. The results are shown in Table \ref{table:Sparse_VS_NonSparse}. Clearly, none of them completely outperforms the other, though CRC$\_l_1$ only wins on 2 of the 13 datasets. Therefore, the question becomes when shall we choose which one if we want to get a better performance, and we expect to get the answer before applying them to the data. For an easy comparison between them, we propose a relative superiority measure called Error Reduction Rate (ERR), which is defined as
\begin{align}\label{eq:ERR}
ERR & = \frac{{Err(CRC\_{l_1}) - Err(CRC\_{l_2})}}{{Err(CRC\_{l_1})}}\\
 & = \frac{{Acc(CRC\_{l_2}) - Acc(CRC\_{l_1})}}{{1 - Acc(CRC\_{l_1})}},
\end{align}
where $Err(\cdot)$ and $Acc(\cdot)$ denote the error rate and accuracy rate of the concerned model, respectively. ERR shows how much performance improvement can be got from replacing CRC$\_l_1$ by CRC$\_l_2$. Therefore, positive ERR values indicate CRC$\_l_2$ performs better than CRC$\_l_1$, while negative ones stand for the opposite. The larger the ERR value is, the more the data favors non-sparse representation.

In order to predict how much a specific dataset might favor non-sparse representation before knowing the ERR value, it is necessary to design some scoring function ($S(l_2, l_1)$) which coincides with ERR. It should be function of some statistics and properties of the dataset. Therefore, we list the representative statistics for each of the 13 datasets as shown in Table \ref{table:Sparse_VS_NonSparse}, including the feature dimensionality $d$, the number of classes $C$, the number of training samples per class $n_i$, and the total number of training samples $n$. Besides that, we believe that datasets should have some properties which are independent from these simple statistics. One of the most important properties is the quality of features, which directly influences the performance of a classification model. An intuitive feeling is that larger ERR may be related to better features, because better features generally enable that samples in the same class stay relatively closer to each other and thus make it easier to generate a discriminative collaborative representation even with non-sparse coefficients. Therefore, we design a Feature Discrimination Rate (FDR) as
\begin{equation}\label{eq:FDR}
    FDR = \frac{{Acc({MPD})}}{{Acc({chance})}},
\end{equation}
where MPD is the simplest Minimum Point-wise Distance based classifier, and ``chance'' is the method of randomly guessing the class label for the test sample(s). In greater details, MPD uses Euclidean distance for point-wise distance measurement and treats the minimum point-wise distance between the test sample(s) and the training samples belong to a specific class as the dissimilarity between them. MPD directly uses such a dissimilarity for classification. Since its performance purely relies on the feature space, it makes FDR a good measure of the features' discriminative power. Note that for FDR, $Acc(\cdot)$ uses the top-1 recognition accuracy, which is different from that for ERR.

\begin{figure}[htb]
\begin{center}
   \includegraphics[width=1.0\linewidth]{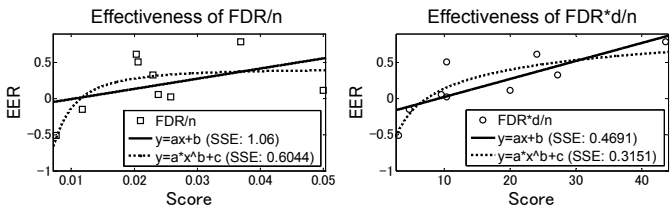}
\end{center}
   \caption{Comparison on the effectiveness of sparse vs. non-sparse representation scoring functions.}
\label{fig:Score_function_comparison}
\end{figure}

Both MPD's accuracy rates and the FDR values for all datasets are given in Table \ref{table:Sparse_VS_NonSparse}. However, FDR itself is not proportional to ERR. Recall that there are many evidences in the literature showing that a larger $d$ can lead to a higher ERR, so we also test the effectiveness of ``FDR$\times d$''. Unfortunately, it still does not coincide with ERR. To enhance it, we propose to have the total number of training samples $n$ involved in an inversely proportional way as
\begin{equation}\label{eq:Score}
    S(l_2, l_1) = \text{FDR} \times d/n.
\end{equation}
The reason is quite simple: a smaller $n$ generally means a lower redundancy of training samples so that the representation coefficients are likely to be denser but not sparser. To prove that, we simply test ``FDR$/n$'' and find it can already differentiate negative ERR values from positive ones. However, we insist that ``FDR$\times d/n$'' is a better choice. When the four special datasets (marked with a star in Table \ref{table:Sparse_VS_NonSparse} because they contain only 1 to 3 trials or have large within-class variations with few samples per class) are not considered, ``FDR$\times d/n$'' clearly has a more definite relationship w.r.t to ERR in terms of a linear or a power function with low SSE (Sum of Squared Errors) than ``FDR$/n$'', as shown in Figure \ref{fig:Score_function_comparison}. Note that the threshold of ``FDR$\times d/n$'' in our experiments (say 5.0) for differentiating positive ERR values from negative ones may not be accurate enough for generalization to other datasets, but we believe that it can be a good reference.



\subsection{Dictionary learning }

We compare the simple dictionary learning model for non-sparse representation (DL-NSCR) with CRC$\_l_2$, CRC$\_l_1$, and the most influential dictionary learning models for sparse representation which are generally more complex than DL-NSCR, including FDDL \cite{FDDL_ICCV11}, LC-KSVD \cite{LC-KSVD_CVPR11}, and DL-COPAR \cite{DL-COPAR_ECCV12}. The results shown in Table \ref{table:DL_Benefits} clearly show that even DL-NSCR can significantly promote CRC$\_l_2$'s performance on 12 of the 13 datasets, and it also outperforms all its competitors on 4 of the 5 classification tasks, except on the person re-identification task. The relatively lower performance boost of DL-NSCR on the re-identification datasets is probably due to the difficulty of the data which limits the room for improvement (all the MPD rates on these datasets are lower than those for the other datasets). Even though, it is interesting to see that DL-NSCR performs the best on the two datasets which favors CRC$\_l_1$ than CRC$\_l_2$. Moreover, when we focus on FDDL as it is most similar to DL-NSCR, we can see that it only wins DL-NSCR on 2 of the 13 datasets (iLIDS-MA23 and iLIDS-MA46). Therefore, generally speaking, when dictionary learning is taken into account, non-sparse representation looks more promising than sparse representation. Note that DL-NSCR is just a simple example, and it will be promising and valuable to explore better dictionary learning models for non-sparse collaborative representation.


\setlength{\tabcolsep}{1.2pt}
\begin{table}[ht]
\caption{Performances of dictionary learning models.}
\centering{ \scriptsize
\begin{tabular}{lcccccccc}
\hline\\[-2ex]
   & &\multicolumn{4}{c}{\textbf{Sparse}} &   &\multicolumn{2}{c}{\textbf{Non-sparse}} \\
              \cline{3-6} \cline{8-9}
              \\[-2.2ex]
\textbf{Dataset} &   & \colorbox[gray]{0.8}{CRC$\_l_1$} & FDDL & LC-KSVD & DL-COPAR & & \colorbox[gray]{0.8}{CRC$\_l_2$} & DL-NSCR \\
\hline \\[-2ex]
Extended Yale B& & 0.951 & 0.968 & 0.944 & 0.928 & & 0.976 & \textbf{0.981} \\
\emph{AR*}     & & 0.898 & 0.917 & 0.677 & 0.694 & & 0.919 & \textbf{0.933} \\
KTH-TIPS       & & 0.957 & 0.699 & 0.885 & 0.585 & & 0.971 & \textbf{0.987} \\
CUReT          & & 0.824 & 0.049 & 0.930 & 0.103 & & 0.933 & \textbf{0.985}\\
Swedish Leaf   & & 0.958 & 0.922 & 0.990 & 0.437 & &\textbf{0.991}&\textbf{0.991}\\
\emph{Food*}   & & 0.311 & 0.175 & 0.220 & 0.167 & & 0.349 & \textbf{0.363} \\
iLIDS-MA10     & & 0.750 & 0.758 & \textbf{0.818} & 0.698 & & 0.778 & \underline{0.770}\\
iLIDS-MA23     & & 0.785 & 0.823 & 0.823 & \textbf{0.853} & & 0.790 & 0.800\\
\emph{iLIDS-MA46*}& &0.800&\textbf{0.850} & 0.825 & \textbf{0.850} & & 0.825 & 0.825\\
iLIDS-AA10     & & 0.658 & 0.655 & \textbf{0.732} & 0.425 & & 0.677 & 0.723\\
iLIDS-AA23     & & 0.731 & 0.740 & 0.747 & 0.472 & & 0.690 & \textbf{0.771} \\
iLIDS-AA46     & & 0.779 & 0.700 & 0.737 & 0.666 & & 0.667 & \textbf{0.792} \\
\emph{CAVIAR4REID*}& &0.478 & 0.476 & \textbf{0.556} & 0.370 & & 0.514 & 0.544 \\
\hline
\end{tabular}}
\label{table:DL_Benefits}
\end{table}
\setlength{\tabcolsep}{5.0pt}
\begin{table*}[ht]
\caption{Running time (minisecond per sample) comparison on four representative datasets. The best results are in bold, while the 2nd best ones are marked in italic.}
\centering{ 
\begin{tabular}{lcccccccc}
\hline\\[-2ex]
   & &\multicolumn{4}{c}{\textbf{Sparse}} &   &\multicolumn{2}{c}{\textbf{Non-sparse}} \\
              \cline{3-6} \cline{8-9}
              \\[-2.2ex]
\textbf{Dataset} &   & \colorbox[gray]{0.8}{CRC$\_l_1$} & FDDL & LC-KSVD & DL-COPAR & & \colorbox[gray]{0.8}{CRC$\_l_2$} & DL-NSCR \\
\hline \\[-2ex]
Ext Yale B& & 0 / 3093 & 2386/1257 & 117 / \textbf{0.56} & 1274 / 18.9 & & 0 / 16.9 & \textbf{83} / \underline{4.4} \\
KTH-TIPS       & & 0 / 2830 & 1138/2371 & 265 / \underline{4.4} & 3154 / 6.9 & & 0 / \textbf{0.18} & \textbf{19.3} / 7.7 \\
Swedish Leaf   & & 0 / 4149 & 1753/3470 & 220 / \underline{3.2} & 1806 / 9.2 & & 0 / \textbf{0.77} & \textbf{191} / 6.0 \\
iLIDS-AA46     & & 0 / 8420 & 134715/10349& 6059 / 21.1 & 539 / 48.7 & & 0 / \underline{9.8} & \textbf{260} / \textbf{8.7} \\
\hline
\end{tabular}}
\label{table:Computational_Cost}
\end{table*}

\subsection{Complexity and running time}

DL-NSCR takes an alternative optimization model which has theoretical guarantee for global convergence and local q-linear (faster than linear) convergence speed \cite{AOConv_NPSC03}. In our experiments, it always converges within several steps. Considering that each iteration only contains basic matrix operations, it is easy to get the complexity for training and testing DL-NSCR. Concretely, the complexity for training is $\mathcal{O}\left((K+n)dKL+K^3L+K^2n\right)$, and that for testing is $\mathcal{O}\left((d+K+L)KL\right)$. It can be seen that DL-NSCR scales less than linearly with $d$, $L$, and $n$, but nearly proportionally to $K^3$. Thus, when the $K$ is predetermined, it scales well with $d$ and $n$. Table \ref{table:Computational_Cost} presents the actual running time for all the concerned methods on four representative datasets. Clearly, DL-NSCR is the fastest dictionary learning model, and its testing time is also comparable to the fastest dictionary learning based sparse representation model.


\section{Conclusion and future work}
\label{sec:conclusions}

We have shown a promising scoring function for pre-selecting sparse or non-sparse collaborative representation models. DL-NSCR, the simple dictionary learning model for non-sparse collaborative representation, has demonstrated its superiority to both sparse and non-sparse collaborative representation models and those state-of-the-art dictionary learning models for sparse representation. There is still a large room for further enhancing it with better dictionary learning models, and it will be very interesting to have a more comprehensive comparison between them and existing dictionary learning models.



%

\appendices

\ifCLASSOPTIONcompsoc
  \section*{Acknowledgments}
\else
  \section*{Acknowledgment}
\fi
This work was supported by ``R\&D Program for Implementation of
Anti-Crime and Anti-Terrorism Technologies for a Safe and Secure
Society'', Funds for integrated promotion of social system reform and
research and development of the Ministry of Education, Culture,
Sports, Science and Technology, the Japanese Government.

\ifCLASSOPTIONcaptionsoff
  \newpage
\fi



\bibliographystyle{IEEEtran}
\bibliography{NSCR_bib}
\end{document}